\documentclass[runningheads]{llncs}
\usepackage{graphicx}
\usepackage{cite}
\begin{document}
\title{Improved Object-Based Style Transfer with Single Deep Network}
%
%
\author{Harshmohan Kulkarni\inst{1} \and
Om Khare\inst{1} \and
Ninad Barve\inst{1} \and
Dr. Sunil Mane\inst{1}}
\authorrunning{H. Kulkarni et al.}
%
\institute{College of Engineering Pune Technological University \\
\email{\{kulkarnihs20.comp, khareom20.comp, barvenm20.comp, sunilbmane.comp\}@coeptech.ac.in}}
\maketitle              
\begin{abstract}
This research paper proposes a novel methodology for image-to-image style transfer on objects utilizing a single deep convolutional neural network. The proposed approach leverages the You Only Look Once version 8 (YOLOv8) segmentation model and the backbone neural network of YOLOv8 for style transfer. The primary objective is to enhance the visual appeal of objects in images by seamlessly transferring artistic styles while preserving the original object characteristics. The proposed approach's novelty lies in combining segmentation and style transfer in a single deep convolutional neural network. This approach omits the need for multiple stages or models, thus resulting in simpler training and deployment of the model for practical applications. The results of this approach are shown on two content images by applying different style images. The paper also demonstrates the ability to apply style transfer on multiple objects in the same image.

\end{abstract}
\section{Introduction}
In the realm of computer vision and image processing, the fusion of advanced segmentation models and image style transfer techniques has opened new avenues for enhancing the visual aesthetics of images. Image style transfer is a technique that combines the content features of one image with the artistic style features of another, creating a new image with the content and appearance of the original image. This is achieved through deep neural networks, typically using convolutional neural networks. The applications of neural style transfer are diverse and extend across various domains. In the realm of art and design, it offers a novel way to generate unique artistic creations, enabling artists to explore different styles and experiment with their visual expressions \cite{art1} \cite{art2} \cite{art3}. Moreover, in the field of entertainment and gaming, NST can be utilized to generate visually stunning graphics and animations, enhancing user experience and immersiveness \cite{ent1} \cite{ent2} \cite{ent3} \cite{ent4}. Furthermore, in the domain of fashion and interior design, NST can aid in generating innovative designs and patterns, facilitating creativity and customization \cite{fas1} \cite{fas2}. Overall, Neural Style Transfer holds promise in revolutionizing various industries by offering a powerful tool for artistic expression, creativity, and visual enhancement.

The existing challenges in image-to-image style transfer primarily revolve around the absence of single deep network models tailored for preserving object characteristics during the stylization process. The task becomes more intricate when applied to segmented objects, as ensuring the faithful representation of these objects while still incorporating artistic styles, requires a sophisticated and integrated approach.

The primary objective of this research is to develop a robust single deep neural network that performs image-to-image style transfer specifically on segmented objects. Leveraging the power of YOLOv8x for accurate and efficient object segmentation, combined with the backbone network of YOLOv8 for style transfer, this single deep network model aims to achieve a product of artistic styles with the distinct features of segmented objects. 

The following sections of the paper are organized as follows: The literature review discusses various approaches used to date for style transfer, object detection, and object-based style transfer. It then identifies the research gaps and the specific findings this study addresses. This is followed by the Proposed Methodology, with subsequent sections detailing the study's results, conclusion, and future work.   

\section{Related Work}
\subsection{Style Transfer}
Style Transfer has been a revolutionary implementation of deep learning in the field of image processing. In recent years, along with Generative AI, Style Transfer is gaining momentum in design and AI Art. It refers to the techniques that process digital images to conform to the appearance or visual style of some other image. The technique finds its origins in texture generation and non photo-realistic rendering (NPR); both of which are now well-established fields. The development of Neural Style Transfer was pioneered by Gatys et. al. \cite{style1}. The paper introduces A Neural Algorithm of Artistic Style, which applies the reference image style features to the content image, returning the stylized content as the output. Yongcheng et. al. \cite{style2} discuss both pre-NST techniques and NST in detail in their review paper. In the paper, Wang et. al. \cite{style3} demonstrate a improved style transfer method. The non-compatible features of standard style capturing models like VGG and deep-learning light ResNet model family is fixed by the introduction of softmax transformation of the activations. This version was named Stylization With Activation smoothinG (SWAG). Human testers preferred the images stylized by SWAG over the normal generations.

\subsection{Object Segmentation}
This paper also proposes to include image segmentation and object detection in the process of style transfer. Image segmentation is the process of analyzing and breaking down an image into regions or segments, each having pixels with similar characteristics. Each segment can then be further used to create filters and masks or be isolated for processing. Object segmentation is its extension where each segment is identified as a unique object. Fully Convolutional Networks, introduced by Long et. al. \cite{fcn} are one of the efficient architectures for semantic segmentation. The U-Net architecture, originally developed for bio-medical segmentation by Ronneberger et. al. \cite{unet}, is derived from FCN. U-Net has been shown to provide precise segmentation with fewer images originally provided for training. Swim Transformer is a novel vision transformer developed by Liu et. al. \cite{swin}, which can be used as a generalized model for multiple image processing tasks. Semantic segmentation using Swin transformers has shown excellent results. Cheng et al. have developed DeepLab, a semantic segmentation model that uses deep convolutional neural networks (DCNNs). It is the latest model used in the PASCAL VOC-2012 Image segmentation task. Segment Anything Model(SAM) developed by Kirillov et. al. \cite{sam} has demonstrated results exceeding all previous fully supervised results. Its implementation by Ultralytics, called Fast SAM, has speeds of 115ms per image. The YOLO (You Only Look Once) family of models is a real-time faster approach in object detection, the latest being Version 8, released at the start of 2023. YOLOv8 derives it's architecture from YOLOv5. The implementation of YOLOv8 by Ultralytics is better than most other available segmentation/detection models across all the parameters. \cite{yolocompare} shows YOLOv8 segmentation model performs better than Mask R-CNN in complex orchard environments.

\subsection{Style Transfer on objects using segmentation}
Various endeavors have been undertaken to integrate both style transfer and image segmentation concepts, particularly for applying style transfer to specific objects. Virtusio et. al. \cite{objectstyle1} have used VGG-19 for style transfer and deep lab-v2 for segmentation module. Psychogyios et. al. \cite{objectstyle2} have used VGG convolutional layers for style transfer, SAM (segment anything model by Facebook) for segmentation, also done border smoothing and support for multiple objects in the paper. Kurzman et. al. \cite{objectstyle3} have introduced class-based style transfer. This technique adds style to a particular class of object after detecting and segmenting it. They have used VGG-16 as the base model for the style transfer module. Castillo et. al. \cite{objectstyle4} use instance aware segmentation method inspired by another paper, style transfer inspired by Gatys et. al. They have used MRF (Markov random fields) to smoothly apply the mask on the segmented part.

\subsection{Research Gaps and Findings}
Thus, based on the aforementioned literature, it is evident that none of the approaches have utilized a singular deep neural network for this specific task. \cite{objectstyle4}, \cite{objectstyle2}, and \cite{objectstyle1} allude to the development of a unified deep network as a potential avenue for future research. Furthermore, enhancements in segmentation models, as employed in the referenced methodologies, remain a focal point for improvement. Leveraging state-of-the-art models such as YOLOv8x-seg for segmentation and the backbone network of YOLOv8 can significantly enhance the segmentation and style transfer process, thereby advancing the effectiveness of the overall approach.

\section{Proposed Methodology}
The proposed framework uses a single deep network of YOLOv8 for both segmentation and style transfer.

\begin{figure*}[ht]
  \centering
  \includegraphics[width=0.99\textwidth]{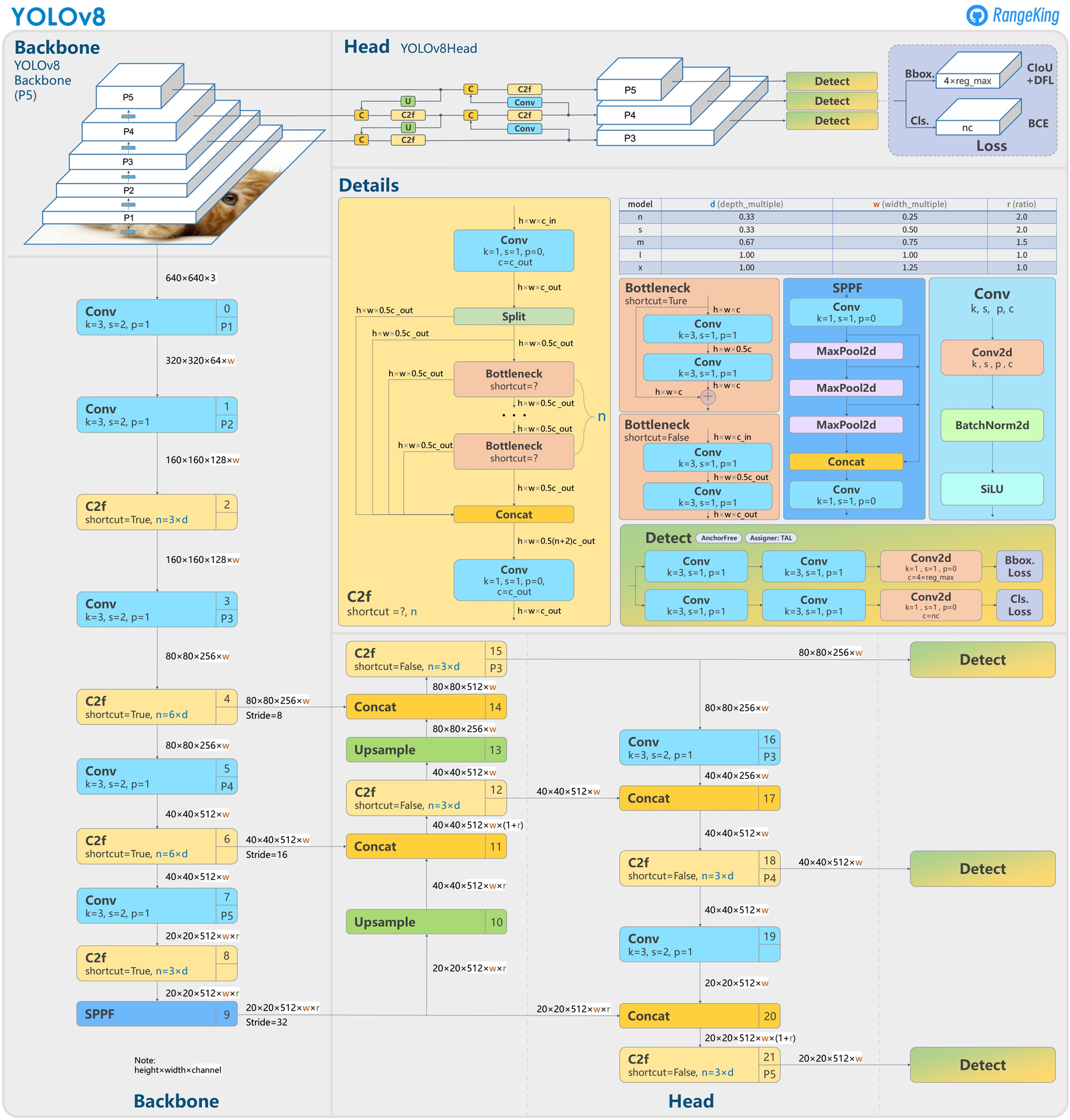}
  \caption{YOLOv8 Architecture, visualization made by GitHub user RangeKing}
  \label{fig:arc}
\end{figure*}

\subsection{YOLOv8 Architecture}

YOLOv8 expands upon the groundwork laid by YOLOv5, incorporating several enhancements in both architecture and developer usability. It introduces a novel backbone network, an innovative loss function, and an advanced anchor-free detection head. In contrast to earlier iterations in the YOLO series, YOLOv8 stands out for its improved speed and precision. This version offers a robust framework for training models across diverse tasks, including object detection, image classification, and instance segmentation.

YOLOv8 introduces significant architectural enhancements as seen in figure \ref{fig:arc}. Notably, it adopts an anchor-free design, distinguishing it from anchor-based models that rely on fixed anchor boxes. Instead of predetermined anchors, YOLOv8 directly predicts object locations and sizes, resulting in more precise object detection.

YOLOv8 architecture can be divided into two parts, as seen in figure \ref{fig:arc}:

\begin{enumerate}
  \item \textbf{Backbone} - The backbone network is responsible for extracting essential features. This crucial component serves as the main feature extraction element of the neural network, aiding in the recognition of objects, differentiation of instances, classification of images downstream, and creation of a map of features obtained.
\\
  \item \textbf{Head} - The detection head utilizes the backbone network features. It plays a pivotal role in predicting object locations and sizes. YOLOv8 introduces an anchor-free detection head, contributing to enhanced accuracy. This innovative component further refines the model's ability to make precise predictions with respect to image segmentation tasks.
\end{enumerate}

Additionally, YOLOv8 introduces a new loss function, crucial for assessing the model's performance during training. This updated loss function compares model predictions to ground truth labels, aiming to minimize the disparity between predictions and actual data. Incorporating this new loss function in YOLOv8 significantly improves its training performance.

\subsection{Architecture Pipeline}
\begin{figure*}[ht]
  \centering
  \includegraphics[width=1\textwidth]{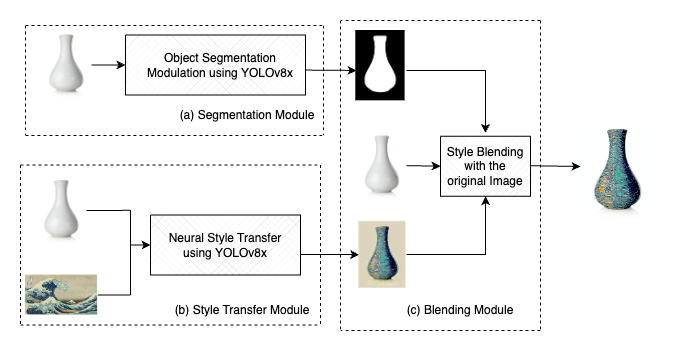}
  \caption{Pipeline Architecture divided into three modules for achieving object-based style transfer using single deep network}
  \label{fig:pipearc}
\end{figure*}

Figure \ref{fig:pipearc} shows the architecture pipeline of the proposed methodology. It can be divided into three modules: Segmentation module, Style Transfer Module, and Blending module.

\subsubsection{Segmentation Module}

This research employs the YOLOv8x-seg model for conducting image segmentation. The YOLOv8-Seg model is an extension of the YOLOv8 object detection model and is designed to perform semantic segmentation in addition to object detection on input images. The model's architecture features a CSPDarknet53 as its backbone for feature extraction, followed by a novel C2f module replacing the conventional YOLO neck module. Subsequent to the C2f module are two segmentation heads. When given an input image, they are responsible for learning and predicting semantic segmentation masks. The model retains a structure akin to YOLOv8 in terms of its detection heads, comprising five detection modules along with a prediction layer.

\subsubsection{Style Transfer Module}

Inspired by [1] Gatys et al., the proposed approach utilizes the internal backbone layers of YOLOv8x-seg model to optimize the losses with the initial pastiche image. For the implementation of the actual style transfer, it minimizes two losses, namely, (1) \textit{Content Loss} and (2) \textit{Style Loss}. The following bullet points will describe more about these losses using \( \delta_l \) as the functional convolutional layer of YOLO network \(l\), \(I\) denotes the pastiche image, \(C\) denotes the content image, \(S\) denotes the style image and \(G(x)\) which is useful for conversion of given input matrix (x) into a gram matrix.

\begin{itemize}
    \item The \textit{Content Loss}, as shown in the Equation \ref{l_content}, for the pastiche image is optimized by using the 7th layer of the Backbone of YOLOv8x-seg network. The target used for optimizing the loss is the convolutional layer output mentioned using the content image \(C\). The formula for the content loss is as follows:
    \begin{equation}
    L_{content}  = (\delta_7(I) - \delta_7(C))^2 \label{l_content}
    \end{equation}
    \item The \textit{Style Loss}, as shown in the Equation \ref{l_style} for the pastiche image is optimized by using a collection of convolutional and Coarse2Fine (C2F) layers from the YOLO's backbone. The gram calculated from the output of style image \(S\) and pastiche image \(P\) from these layers are used for the optimization. The loss calculated at each layer is multiplied by a weight factor \(W_l\), highlighting the significance of certain layers over others in the overall loss. The formula for the content loss is as follows:
    \begin{equation}
    L_{style}  = \sum_{l}W_l(G(\delta_l(S)) - G(\delta_l(P)))^2 \label{l_style}
    \end{equation}
\end{itemize}

As shown in equation \ref{l_total}, \(L_{total}\) is the total loss calculated by a weighted sum of the losses due to content and style transfer, with hyperparameters of \(\alpha\) and \(\beta\) determining the importance of each of them. The equation for this is as follows: 
    \begin{equation}
    L_{total} = \alpha L_{content} + \beta L_{style} \label{l_total}
    \end{equation}

The overall objective of this module is to find the optimal generated image that minimizes \(L_{total}\), thus achieving a harmonious blend of content and style in the final composition.

\subsubsection{Blending Module}
This module receives the stylized image generated by style transfer along with the corresponding mask delineating the targeted object for stylization. Then, the mask is used to isolate the object from the stylized image. The process of isolation includes pixel-wise multiplication of the styled image with the mask. By this process, only the pixels corresponding to the styled object will remain in the image, while the rest will be marked out. Then, the isolated object is integrated back into the original image by adding the styled portion back into the original image, completing the object-based neural style transfer.

\section{Results}
The ensuing section presents the outcomes achieved by implementing our proposed methodology. These results provide a comprehensive overview of the effectiveness and capabilities of our approach.

\subsection{Results for object-based style transfer using YOLOv8 Image Segmentation Model on single object}

\begin{figure*}[ht]
  \centering
  \includegraphics[width=1\textwidth]{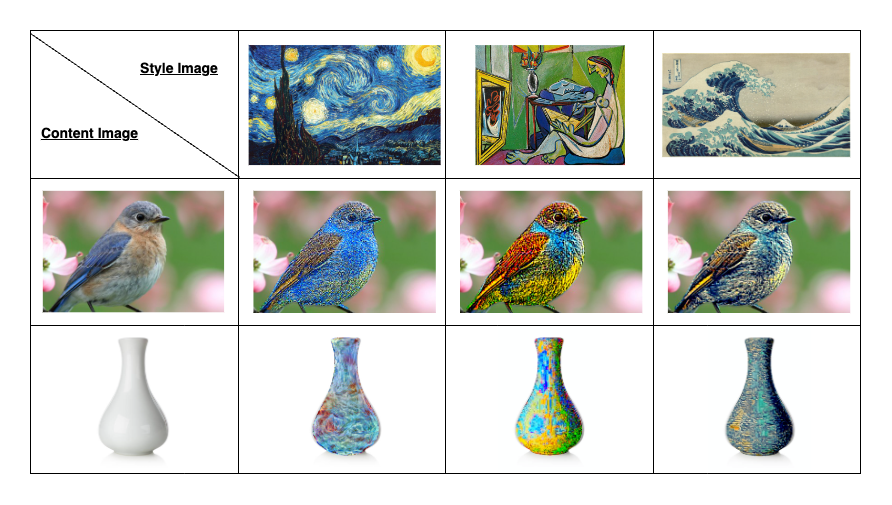}
  \caption{Results for object-based style transfer using YOLOv8 Image Segmentation Model on single object}
  \label{fig:results1}
\end{figure*}

\begin{figure*}[ht]
  \centering
  \includegraphics[width=1\textwidth]{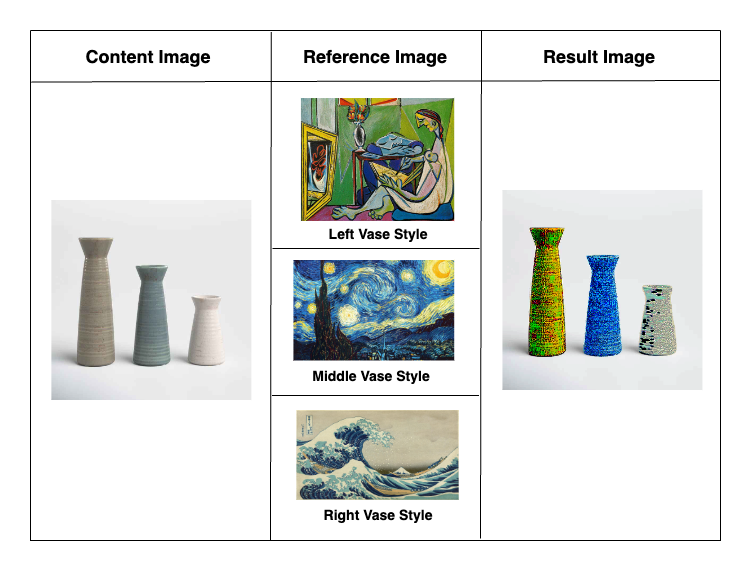}
  \caption{Results for object-based style transfer using multiple style images on multiple objects within a single image}
  \label{fig:results2}
\end{figure*}

The proposed methodology demonstrates its ability to create visually captivating images by applying the style features of three iconic paintings: The Starry Night by Vincent Van Gogh, La Muse by Pablo Picasso, and The Great Wave off Kanagawa by Hokusai. These styles are seamlessly transferred onto images featuring a bird and a vase, as shown in Figure \ref{fig:results1}. This figure highlights our approach's effectiveness and remarkable accuracy in translating the essence of these paintings into object images. Notably, the preservation of the original object image's content is evident in the intricate details of the feathers and wings in the generated bird image, as well as in the visually compelling rendition of the vase through the style transfer. This preservation of original features ensures that while the objects adopt the artistic styles of the chosen paintings, their essential characteristics remain intact, resulting in visually appealing compositions.

\subsection{Results for object-based style transfer using multiple style images on multiple objects within a single image}

Figure \ref{fig:results2} illustrates the outcomes achieved through object-based style transfer on multiple objects within a single image, utilizing the same three style images mentioned earlier. The image reveals the presence of three vases in the content image. Our proposed approach effectively detects these distinct objects and applies three different styles to each vase while maintaining the original texture of each object. Specifically, the style features of La Muse are seamlessly transferred to the left vase, The Starry Night's features to the middle vase, and The Great Wave's features to the right vase. This precise application of varied styles to different objects within the same image demonstrates the versatility and effectiveness of our methodology in creating diverse and visually engaging compositions.

\section{Conclusion and Future Work}

Style Transfer models on their own transform the entire content image based on the style image provided. Introducing Object Detection improves the locality of the style. Improving on the two model implementations of the papers cited in the Literature review, The proposed approach implements a single Deep Convolutional Neural Network for Object Detection based Style Transfer. The proposed approach implements the latest YOLOv8x segmentation model and has modified the architecture to introduce style transfer capabilities. The results have demonstrated that it is possible to combine both tasks of Object Detection and Style Transfer into a single model.

Future work involves delving into additional filters to retain the distinct traits of both the content and style images. This could enhance the controllability of style transfer by offering more nuanced ways to integrate style elements into the content. Another potential avenue for advancement lies in extending this style transfer method to 3D objects, thereby broadening its application beyond traditional 2D images and unlocking new possibilities in immersive environments and augmented reality experiences.

\bibliographystyle{splncs04}
\bibliography{references.bib}

\begin{thebibliography}{10}
\providecommand{\url}[1]{\texttt{#1}}
\providecommand{\urlprefix}{URL }
\providecommand{\doi}[1]{https://doi.org/#1}

\bibitem{fas2}
Alcaraz, L.M., Hu, R., Agrawal, A.: Generating fashion through neural style transfer

\bibitem{objectstyle4}
Castillo, C., De, S., Han, X., Singh, B., Yadav, A.K., Goldstein, T.: Son of zorn's lemma: Targeted style transfer using instance-aware semantic segmentation. In: 2017 IEEE International Conference on Acoustics, Speech and Signal Processing (ICASSP). pp. 1348--1352. IEEE (2017)

\bibitem{ent3}
Deniz, {\c{S}}., K{\"U}{\c{C}}{\"U}KKAYKI, H.T., S{\"u}rer, E.: Automated game mechanics and aesthetics generation using neural style transfer in 2d video games. Bili{\c{s}}im Teknolojileri Dergisi  \textbf{14}(3),  287--300 (2021)

\bibitem{fas1}
Ganesan, A., Oates, T., et~al.: Fashioning with networks: Neural style transfer to design clothes. arXiv preprint arXiv:1707.09899  (2017)

\bibitem{art1}
Gatys, L.A., Ecker, A.S., Bethge, M.: A neural algorithm of artistic style. arXiv preprint arXiv:1508.06576  (2015)

\bibitem{style1}
Gatys, L.A., Ecker, A.S., Bethge, M.: Image style transfer using convolutional neural networks. In: Proceedings of the IEEE Conference on Computer Vision and Pattern Recognition (CVPR) (June 2016)

\bibitem{ent2}
He, J.: Exploring style transfer algorithms in animation: Enhancing visual. Entertainment Computing  \textbf{49},  100625 (2024)

\bibitem{ent1}
Huang, H., Wang, H., Luo, W., Ma, L., Jiang, W., Zhu, X., Li, Z., Liu, W.: Real-time neural style transfer for videos. In: Proceedings of the IEEE conference on computer vision and pattern recognition. pp. 783--791 (2017)

\bibitem{ent4}
Ioannou, E., Maddock, S.: Towards real-time g-buffer-guided style transfer in computer games. IEEE Transactions on Games  (2024)

\bibitem{style2}
Jing, Y., Yang, Y., Feng, Z., Ye, J., Yu, Y., Song, M.: Neural style transfer: A review. IEEE Transactions on Visualization and Computer Graphics  \textbf{26}(11),  3365--3385 (2020). \doi{10.1109/TVCG.2019.2921336}

\bibitem{sam}
Kirillov, A., Mintun, E., Ravi, N., Mao, H., Rolland, C., Gustafson, L., Xiao, T., Whitehead, S., Berg, A.C., Lo, W.Y., Dollar, P., Girshick, R.: Segment anything. In: Proceedings of the IEEE/CVF International Conference on Computer Vision (ICCV). pp. 4015--4026 (October 2023)

\bibitem{objectstyle3}
Kurzman, L., Vazquez, D., Laradji, I.: Class-based styling: Real-time localized style transfer with semantic segmentation. In: Proceedings of the IEEE/CVF International Conference on Computer Vision Workshops. pp.~0--0 (2019)

\bibitem{swin}
Liu, Z., Lin, Y., Cao, Y., Hu, H., Wei, Y., Zhang, Z., Lin, S., Guo, B.: Swin transformer: Hierarchical vision transformer using shifted windows. In: Proceedings of the IEEE/CVF international conference on computer vision. pp. 10012--10022 (2021)

\bibitem{fcn}
Long, J., Shelhamer, E., Darrell, T.: Fully convolutional networks for semantic segmentation. In: Proceedings of the IEEE Conference on Computer Vision and Pattern Recognition (CVPR) (June 2015)

\bibitem{objectstyle2}
Psychogyios, K., Leligou, H.C., Melissari, F., Bourou, S., Anastasakis, Z., Zahariadis, T.: Samstyler: Enhancing visual creativity with neural style transfer and segment anything model (sam). IEEE Access  (2023)

\bibitem{yolocompare}
Ranjan~Sapkota, D.A., Karkee, M.: Comparing yolov8 and mask rcnn for object segmentation in complex orchard environments

\bibitem{unet}
Ronneberger, O., Fischer, P., Brox, T.: U-net: Convolutional networks for biomedical image segmentation. In: Medical Image Computing and Computer-Assisted Intervention--MICCAI 2015: 18th International Conference, Munich, Germany, October 5-9, 2015, Proceedings, Part III 18. pp. 234--241. Springer (2015)

\bibitem{art2}
Ruder, M., Dosovitskiy, A., Brox, T.: Artistic style transfer for videos. In: Pattern Recognition: 38th German Conference, GCPR 2016, Hannover, Germany, September 12-15, 2016, Proceedings 38. pp. 26--36. Springer (2016)

\bibitem{art3}
Ruder, M., Dosovitskiy, A., Brox, T.: Artistic style transfer for videos and spherical images. International Journal of Computer Vision  \textbf{126}(11),  1199--1219 (2018)

\bibitem{objectstyle1}
Virtusio, J.J., Talavera, A., Tan, D.S., Hua, K.L., Azcarraga, A.: Interactive style transfer: Towards styling user-specified object. In: 2018 IEEE Visual Communications and Image Processing (VCIP). pp.~1--4. IEEE (2018)

\bibitem{style3}
Wang, P., Li, Y., Vasconcelos, N.: Rethinking and improving the robustness of image style transfer. In: Proceedings of the IEEE/CVF Conference on Computer Vision and Pattern Recognition (CVPR). pp. 124--133 (June 2021)

\end{thebibliography}
\end{document}